\pdfoutput=1

\documentclass[11pt]{article}

\usepackage[]{acl}

\usepackage{times}
\usepackage{latexsym}

\usepackage[T1]{fontenc}

\usepackage[utf8]{inputenc}

\usepackage{microtype}

\usepackage{amsmath}            
\usepackage{amsfonts}           
\usepackage{amssymb}

\usepackage{subcaption}  
\usepackage{graphicx}
\usepackage{adjustbox}
\usepackage{amsthm}
\usepackage{fancyhdr}

\usepackage{verbatim}           
\usepackage{lastpage}

\usepackage{fancyhdr}
\usepackage{float}

\usepackage{longtable}

\usepackage{booktabs}
\usepackage{siunitx}

\usepackage{natbib}
\usepackage{url}
\usepackage{algorithm2e}
\usepackage{color}
\usepackage{longtable}
\usepackage[normalem]{ulem}
\usepackage{comment}
\usepackage{fancyvrb}
\usepackage{todonotes}

\newcommand*{\affaddr}[1]{#1} 
\newcommand*{\affmark}[1][*]{\textsuperscript{#1}}
\newcommand*{\email}[1]{\texttt{#1}}

\def\JNdel#1{\bgroup\markoverwith{\textcolor{red}{\rule[0.5ex]{2pt}{1pt}}}\ULon{#1}}

\title{Data-driven Approach to Differentiating between Depression and Dementia from Noisy Speech and Language Data}

\author{%
Malikeh Ehghaghi\affmark[1,2], Frank Rudzicz\affmark[1,2,3,4,5], Jekaterina Novikova\affmark[1]\\
\affaddr{\affmark[1]Winterlight Labs, Toronto, ON}\\
\affaddr{\affmark[2]Department of Computer Science, University of Toronto, ON}\\
\affaddr{\affmark[3]Vector Institute for Artificial Intelligence, Toronto, ON}\\
\affaddr{\affmark[4]Li Ka Shing Knowledge Institute, St Michael’s Hospital, Toronto, ON}\\
\affaddr{\affmark[5]Surgical Safety Technologies Inc., Toronto, ON}\\
\email{\{malikeh,jekaterina\}@winterlightlabs.com,\{frank\}@spoclab.com}\\
}

\begin{document}
\maketitle
\begin{abstract}
A significant number of studies apply acoustic and linguistic characteristics of human speech as prominent markers of dementia and depression. However, studies on discriminating depression from dementia are rare. Co-morbid depression is frequent in dementia and these clinical conditions share many overlapping symptoms, but the ability to distinguish between depression and dementia is essential as depression is often curable. In this work, we investigate the ability of clustering approaches in distinguishing between depression and dementia from human speech. We introduce a novel aggregated dataset, which combines narrative speech data from multiple conditions, i.e., Alzheimer's disease, mild cognitive impairment, healthy control, and depression. We compare linear and non-linear clustering approaches and show that non-linear clustering techniques distinguish better between distinct disease clusters. Our interpretability analysis shows that the main differentiating symptoms between dementia and depression are acoustic abnormality, repetitiveness (or circularity) of speech, word finding difficulty, coherence impairment, and differences in lexical complexity and richness. 

\end{abstract}
\section{Introduction}
\label{sec:intro}
Depressive disorder and dementia are clinical conditions that both impose a substantial cost globally in terms of mortality and morbidity and have a significant negative impact on social and economic productivity \citep{jaeschke2021global}. Distinguishing between these conditions has proven to be a challenging task \cite{murray2010distinguishing} as they frequently co-occur and have many overlapping symptoms such as apathy \cite{lee2003depression}, changes in sleep patterns \citep{thorpe2009depression}, and concentration issues \cite{korczyn2009depression}.  However, depression is generally curable by either psychotherapy or medication, while dementia is a neurodegenerative disease, which is caused by irreversible deterioration of the nervous system.  It is hence crucial to differentiate between these two conditions \citep{fraser2016detecting}.

Previous studies demonstrated that machine learning methods and speech analysis are useful in detecting dementia from depression \cite{fraser2016detecting, murray2010distinguishing}. However, the machine learning methods used in prior studies suffer from three main limitations:

Firstly, the datasets applied in prior literature only comprise Alzheimer's disease (AD), healthy control (HC), and depression (Depr) samples of senior participants with similar demographic distributions and recording environments \cite{fraser2016detecting, murray2010distinguishing}. In real world settings, the datasets are very noisy due to variations in the data collection procedures. Additionally, dementia is not necessarily of the AD type in all cases, and other types of dementia like mild cognitive impairment (MCI) can be included. 

Secondly, to the best of our knowledge, previous studies have only used classification approaches to detect AD from HC \cite{pulido2020alzheimer,balagopalan2021comparing,balagopalan2021bcomparing}, Depr from HC \cite{wu2022automatic}, or AD from Depr \cite{fraser2016detecting} using speech.  This might not be an ideal simulation of the real world diagnosis procedure. In clinical diagnosis, the first step is to detect the symptoms and explore the pattern changes in patient records before diagnosing the disease \cite{regier2013dsm}, while in classification, we first map the samples to the disease labels and then, apply interpretability methods to explore the differentiating features between the classes \cite{gordon1999classification}.

Lastly, prior studies demonstrated that acoustic and linguistic features extracted from spontaneous speech provide valuable indicators of both mental disorders such as depression \citep{low2020automated} and cognitive impairment like AD or MCI \citep{fraser2016linguistic,boschi2017connected}. However, they did not derive a strong conclusion about the main distinguishing speech-based symptoms in classifying dementia from depression \cite{fraser2016detecting}.

To address the first limitation, we generate a novel aggregated dataset, which combines several speech datasets comprising AD, MCI, HC, and Depr labels with a variety of data collection procedures.  To address the second and third limitations, we introduce a novel approach, which applies clustering techniques to inspect what data-driven feature categories (symptoms) are the main differentiators between AD, MCI, Depr, and HC samples. We then use the distinguishing symptoms as a feature selection technique to classify AD, MCI, and Depr. Our key findings indicate that 1) the non-linear clustering approaches outperform the linear techniques in terms of separability level of distinct disease clusters; 2) acoustic abnormalities, variations in lexical complexity and richness, repetitiveness (or circularity) of speech, word finding difficulty, and coherence impairment are the main differentiating symptoms to distinguish between different types of dementia (e.g., AD and MCI), and Depr; 3) data-driven differentiators are able to substantially improve performance of classification across diseases.

\section{Related Work}
\label{sec:background} 
There has been a substantial number of studies on detecting either dementia (e.g., MCI or AD) or depression from spontaneous speech. However, little has been done to distinguish dementia from depression using discourse patterns. 

To discriminate dementia from depression, \citet{fraser2016detecting} applied speech data from the Pitt corpus in the DementiaBank database \cite{becker1994natural}, elicited from elderly participants through picture description task, with `Cookie Theft' \cite{goodglass2001bdae} used as a picture. The samples were labeled as either AD or HC based on a personal history and a neuropsychological assessment battery \cite{iverson2008neuropsychological}. A subset of the samples were labeled as depressed or non-depressed based on the established threshold on Hamilton Depression Rating Scale (HAM-D) test scores \cite{bagby2004hamilton}. To explore the distinguishing discourse patterns between AD and Depr, \citet{murray2010distinguishing} collected a speech dataset of elderly participants (with Depr, AD, or HC labels) who completed a picture description task, with Norman Rockwell’s painting `The Soldier' used as a picture. Samples with Depr were diagnosed based on DSM-IV criteria \cite{frances1995dsm} and samples with AD met NINCDS-ADRDA criteria \cite{tierney1988nincds} for probable AD. The datasets used in these studies didn't include other types of dementia such as MCI, and all of their samples followed the same data collection procedure, while we create an aggregated dataset, which consists of AD, MCI, HC, and Depr samples from different speech datasets with various data collection procedures.

\citet{murray2010distinguishing} examined whether elderly individuals with depression can be distinguished from those at early stages of AD through distinct patterns in narrative speech. Based on their findings, individuals with AD generated less informative speech compared to the depressed patients in their picture descriptions, while there were no significant differences in the informativeness of the narratives between HC and Depr samples. Furthermore, quantitative and syntactic measures of discourse did not differ across the three groups. However, \citet{murray2010distinguishing} did not attempt to make predictions using the data.

\citet{fraser2016detecting} investigated if the automated AD screening tools misclassify cognitively healthy
participants with Depr as AD when using narrative speech.  They also used linguistic and acoustic features to classify non-depressed AD subjects from those with comorbid depression from speech elicited through picture description task. In their study, they compared logistic regression (LR) with support vector machines (SVM) classification models. Their performance in distinguishing between depressed and non-depressed AD samples was moderate (accuracy = 0.658) due to a wide range of overlapping symptoms. In addition, they only applied classification approaches and they didn't derive the most informative features discriminating between AD patients with and without depression. In the present work, we apply clustering approaches to cluster the diseases based on the similarities in the discourse patterns, and apply interpretability techniques to explore the distinguishing feature categories (symptoms) between distinct diagnosis labels (i.e., HC, AD, MCI, and Depr). We use the differentiating symptoms as a feature selection technique to classify the diseases.

\section{Methods}
\subsection{Dataset}
\label{sec:data-aggregation}
In this paper, we generated an aggregated superset of the datasets listed in Table \ref{tab:speech-datasets} that contains speech recordings of English-speaking participants describing pictures. All the audio recordings were manually transcribed by trained transcriptionists, using the CHAT protocol and annotations \citep{macwhinney2014childes}.

\begin{table}[htbp]
\centering
  \resizebox{\columnwidth}{!}{
      \begin{tabular}{lllll}
          \hline
          \bfseries Dataset & \bfseries AD & \bfseries MCI & \bfseries Depr & \bfseries HC\\
          \hline
          DementiaBank \citep{becker1994natural} & 178 &  138 & 0 & 229\\
          Healthy Aging & 0 & 214 & 0 & 211\\
          ADReSS \citep{luz2020alzheimer} & 54 & 0 & 0 & 54\\
          DEPAC+ \citep{tasnim2022depac} & 0 & 0 & 222 & 532\\
          AD Clinical Trial  & 1616 & 0 & 0 & 0\\ \hline \hline
          Aggregated dataset & 1848 & 352 & 222 & 1026 \\
          \hline 
          
      \end{tabular}
  }
  \caption{Speech datasets used. For each dataset, the number of samples with each diagnosis label is reported in the following columns.}
  \label{tab:speech-datasets}
\end{table}

\textbf{DementiaBank} \citep{becker1994natural} and \textbf{ADReSS} \citep{luz2020alzheimer} are the datasets of pathological speech elicited from participants through picture description task, with `Cookie Theft' \citep{goodglass2001bdae} used as a picture. The recordings are labeled as AD, MCI, and HC.

\textbf{Healthy Aging} is the dataset of speech elicited from  community volunteers through picture description task, with `Family in the Kitchen', `Man in the Living Room', `Food Market', `Picnic', `Grandmother's Birthday', and `Romantic Dinner' proprietary images. The recordings are labeled as possible HC and MCI. Soft labels are based on the established threshold on Montreal Cognitive Assessment \citep{nasreddine2005montreal} screening tool.

\textbf{DEPAC+} is the extended version of the \textbf{DEPAC} \citep{tasnim2022depac} dataset, with more samples collected using the same data collection procedure. This is a dataset of narrative speech elicited from participants through picture description task, with `Family in the Kitchen' and `Man Falling' images. The recordings are labeled as HC and Depr. Soft labels are based on the established threshold on Patient Health Questionnaire-9 (PHQ-9) \citep{kroenke2001phq} test scores\footnote{The participants with a PHQ-9 score \(\leq9\) were labeled as HC, and the remaining samples with a PHQ-9 score \(\geq10\) met criteria for symptoms of depression.}.

\textbf{AD Clinical Trial} is a dataset of speech recordings from the baseline and screening visits of a clinical trial elicited from participants through picture description task, with `Family in the kitchen', `Man in the Living Room', `Grandmother's Birthday', `Romantic Dinner', and `Cookie Theft' \citep{goodglass2001bdae} images. All the recordings are labeled as AD according to the the National Institute on Aging/Alzheimer's Association citeria \citep{frisoni2011revised}.

All images other than `Cookie Theft' \citep{goodglass2001bdae} were designed to match the `Cookie theft' picture in style and the amount of information content units according to picture design principles described by \citet{patel2014park}.

\subsection{Feature Extraction}
\label{sec:feature-extraction}
We extracted 220 acoustic features from audio, and 325 linguistic features from the associated transcripts. These features were classified into the following  categories (the full list is in Appendix \ref{apd:apd_features}):

\textbf{Acoustic: }This category includes spectral and voicing-related features (e.g., Mel-Frequency Cepstral Coefficients (MFCC) \citep{rudzicz2012vocal}, Fundamental frequency $(F_0)$, or statistical functionals of Zero-Crossing Rate (ZCR) \citep{kulkarni2018use}) describing the acoustic properties of the sound wave.

\textbf{Syntactic Complexity: }This category comprises variables like the frequencies  of  various  production  rules from the constituency parsing tree of the transcripts \citep{chae2009predicting}, or Lu’s syntactic complexity features \citep{lu2010automatic} enumerating the rate of usage of different syntactic structures.

\textbf{Discourse Mapping: }This category consists of features such as utterance distances, or speech-graph features \citep{mota2012speech} like graph density \citep{mirheidari2018detecting} to calculate the repetitiveness or circularity of speech.

\textbf{Lexical Complexity and Richness: }This category accounts for the variables like frequency of words, or measures of vocabulary diversity such as type-token ratio \citep{richards1987type} describing the lexical complexity and vocabulary richness of the transcripts.

\textbf{Information Content Units: }This category includes variables such as the number of objects, subjects, locations, and actions used to measure the number of items correctly named in the picture description task previously found to be associated with memory impairment \citep{croisile1996comparative}.

\textbf{Sentiment: }This category contains features such as valence, arousal, and dominance scores \citep{warriner2013norms} for all words and word types describing the sentiment of the words used.

\textbf{Word Finding Difficulty: }This category consists of features including speech rate, duration of words, and number of filled (e.g., um, uh) and unfilled pauses as signs of word finding difficulty, which result in less fluid or fluent speech. 

\textbf{Coherence (Global and Local): } This category includes variables like average, minimum, and maximum cosine distances \citep{mirheidari2018detecting} between subsequent utterances (local coherence) or between utterances and key words (global coherence) using word2vec \citep{church2017word2vec} representation of the utterances to calculate their semantic similarity.

\section{Proposed Novel Approach: Data-Driven Approach to Detecting Differentiating Speech-based Symptoms between Dementia and Depression}
\subsection{Dimensionality Reduction and Clustering}
 We first applied dimensionality reduction techniques to the preprocessed features (see Appendix \ref{apd:data-preprocessing}). To explore linear dimension reduction approaches, we applied Principal Component Analysis (PCA) \citep{wold1987principal} as well as  Linear Discriminant Analysis (LDA) \citep{izenman2013linear}. For non-linear dimensionality reduction techniques, we used Uniform Manifold
Approximation and Projection (UMAP) \citep{mcinnes2018umap} and  T-distributed Stochastic Neighbour Embedding (t-SNE) \citep{van2008visualizing} (See details of implementation and hyperparameter setting in Appendix \ref{apd:param-setting}). 

Next, we clustered the low-dimensional data points by K-Means clustering \citep{mysiak2020explaining} to group them in an unsupervised way into distinguishable clusters. Clusters were meant to represent groups associated with data labels - HC, AD, MCI, and Depr.

\subsubsection{Performance Metrics}
\label{sec:performance-metrics}

The performance of the clustering methods was measured based on the following metrics:

\begin{enumerate}
  \item \textit{Optimal number of disease clusters} determined by the elbow method \citep{yuan2019research}) after training K-Means clustering on the feature embeddings resulted from dimension reduction. The ideal case is to derive 4 distinct disease clusters in line with the 4 diagnosis labels in the aggregated dataset (i.e., HC, AD, MCI, and Depr).
  \item \textit{Silhouette score} \citep{rousseeuw1987silhouettes} was used to measure the level of cluster separability. Its value ranges from -1 to 1. `1' means clusters are well apart from each other and clearly distinguished. `0' means that the distance between clusters is not significant. `-1' means clusters are assigned in the wrong way  \citep{bhardwaj2020silhouette}. The results were recorded for K=4 (the number of labels in the dataset), where K is the number of clusters generated by K-Means clustering.
\end{enumerate}

\subsection{Analysis of the Differentiating Feature Categories between the Disease Clusters}
\label{sec:analyze-feat-cat}
Analysis of the differentiating feature categories across the disease clusters consists of 3 main steps: LIME-based explanation of the low-dimentional embeddings, analysis of feature contributions to the non-linear components, and feature selection using the differentiating feature categories in classification of AD vs MCI vs Depr.

\subsubsection{Local Explanation of the Non-linear Embeddings by LIME}
\label{sec:lime-for-tsne}
We applied a LIME-for-t-SNE\footnote{Publicly available at \url{https://github.com/vu-minh/mlteam-lime-for-tsne}} interpretability method developed by \citealp{bibal2020explaining} to find the main differentiating feature categories between AD, Depr, MCI, and HC diagnosis labels. This approach adapts Local Interpretable Model-agnostic
Explanations (LIME) \citep{ribeiro2016should} to locally explain t-SNE components.

\subsubsection{Analysis of Feature Contributions to the Non-linear Components}
\label{sec:feat-contribution}
\begin{figure}[t]
    \centering
    \begin{subfigure}{0.48\textwidth}
    \centering
        \includegraphics[width=\linewidth]{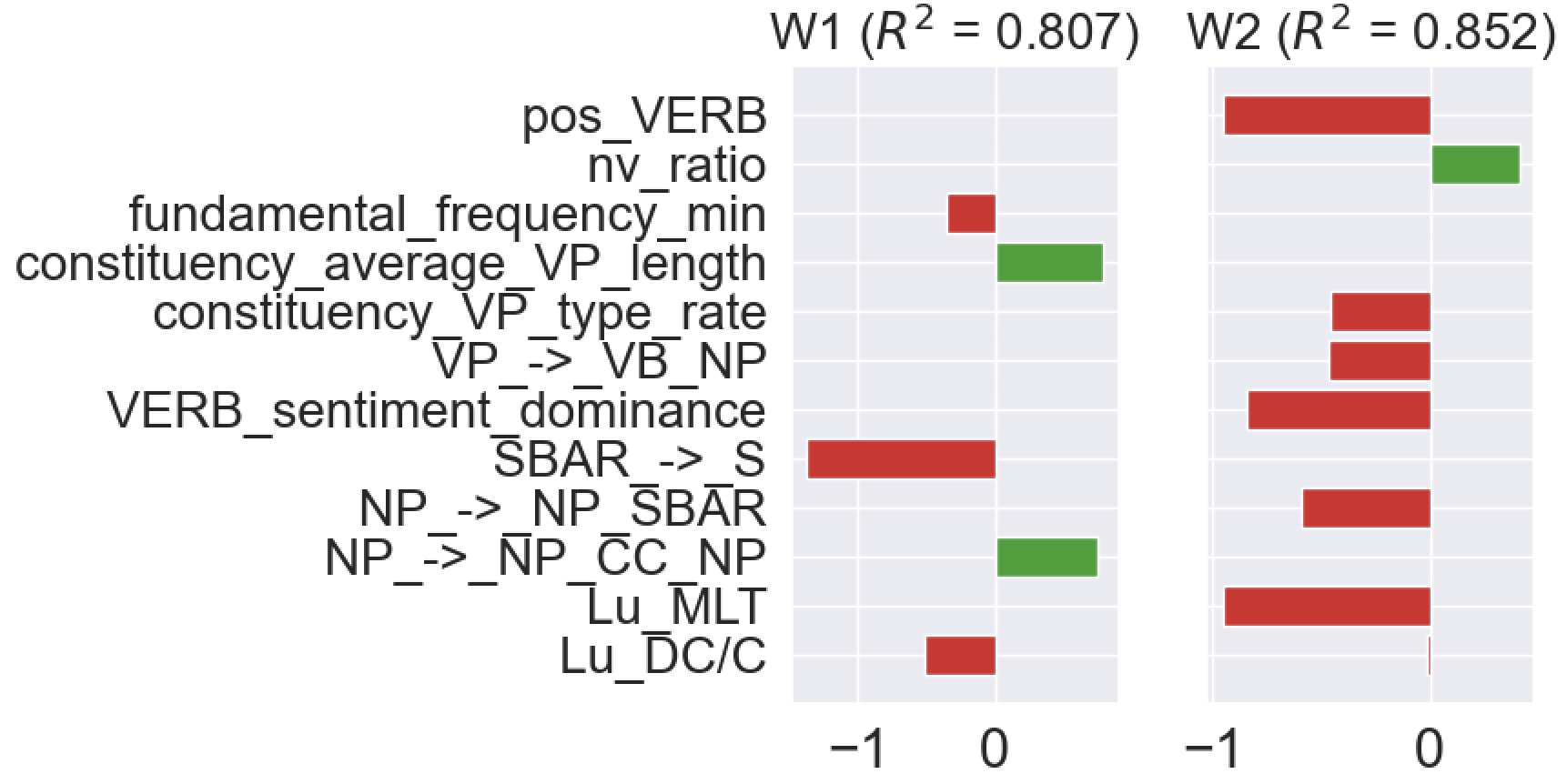}
        \label{fig:lime-weights}
    \end{subfigure}%
    \quad
    \begin{subfigure}{0.48\textwidth}
    \centering
        \includegraphics[width=\linewidth]{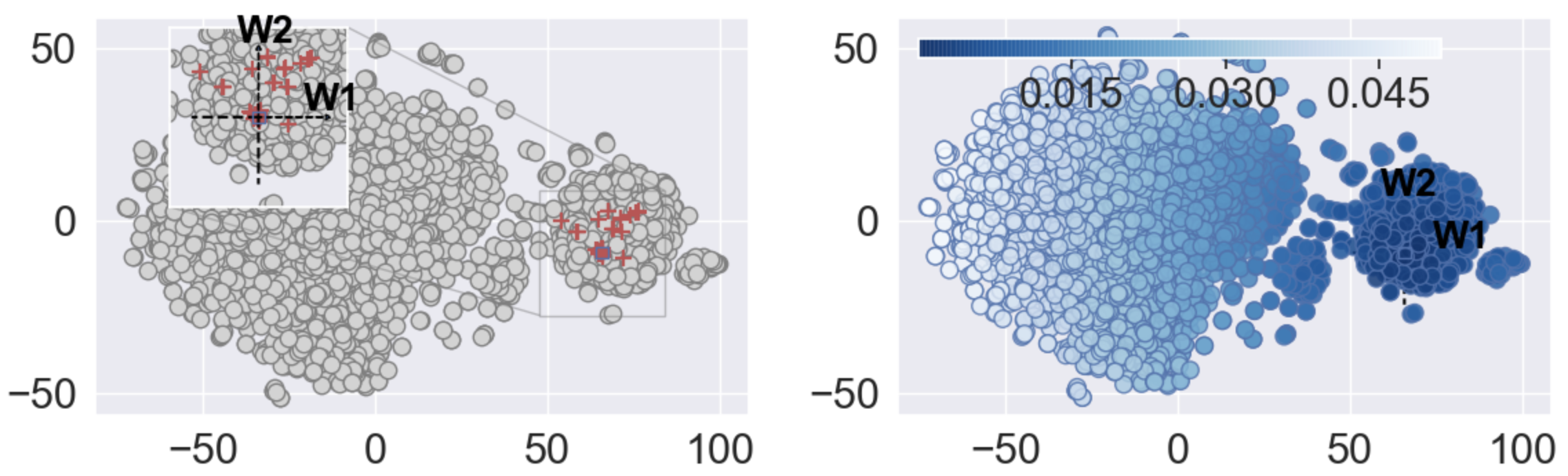}
        \label{fig:lime-clusters}
    \end{subfigure}
    \caption{Explanation of the local trends in the t-SNE embeddings for a selected Depr instance. The figure at the top indicates the weights of the highly-contributed features explaining each local dimension ($R^2$ score indicates how well the local trends are linearly explained per each axis.) The blue transparency in the scatter plot represents the errors of the linear model applied locally on the original instance. The figure at the bottom left is a zoom on the zone of interest for local explanation, with projected samples in red \cite{bibal2020explaining}}
    \label{fig:lime-for-tsne}
\end{figure}

\begin{table*}[t]
    \centering
    \scriptsize
    \begin{tabular}{lcc}
    \hline
    \textbf{\begin{tabular}[c]{@{}l@{}}Dimension reduction\\ method\end{tabular}} &
      \textbf{\begin{tabular}[c]{@{}c@{}}Is the optimal number of\\ clusters (K) equal to 4? \end{tabular}} &
      \textbf{\begin{tabular}[c]{@{}c@{}}Silhouette score \\ (K = 4)\end{tabular}} \\ \hline \hline
    PCA (linear)        & -           & 0.2010\\ 
    LDA (linear)        & -           & 0.4125\\ \hline
    t-SNE (non-linear)  & \textbf{x}  & 0.4723\\ 
    UMAP (non-linear)   & \textbf{x}  & \textbf{0.5580} \\ \hline
    \end{tabular}
    \caption{Summary of the performance of all dimensionality reduction techniques. The second column checks if the optimal number of clusters is equal with the total number of labels (e.g., HC, MCI, AD, and Depr) in the aggregated dataset. `K' refers to the number of clusters in K-Means clustering applied on the embeddings in the low-dimensional space.}
    \label{tab:summary-table}
\end{table*}

\begin{figure*}[t]
    \centering
    \begin{subfigure}{0.4\textwidth}
    \centering
        \includegraphics[width=5.5cm]{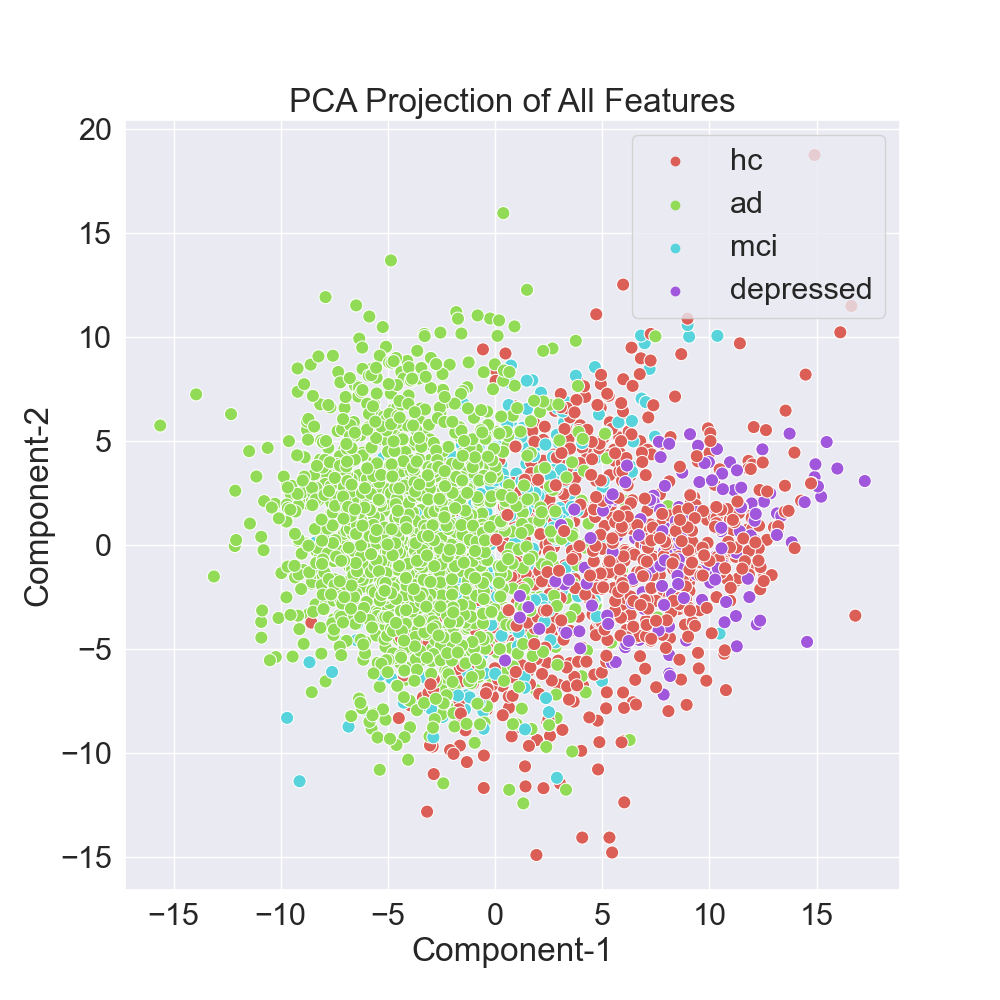}
        \caption{PCA - Disease Clusters}
        \label{fig:pca-disease}
    \end{subfigure}%
    \quad
    \begin{subfigure}{0.4\textwidth}
    \centering
        \includegraphics[width=5.5cm]{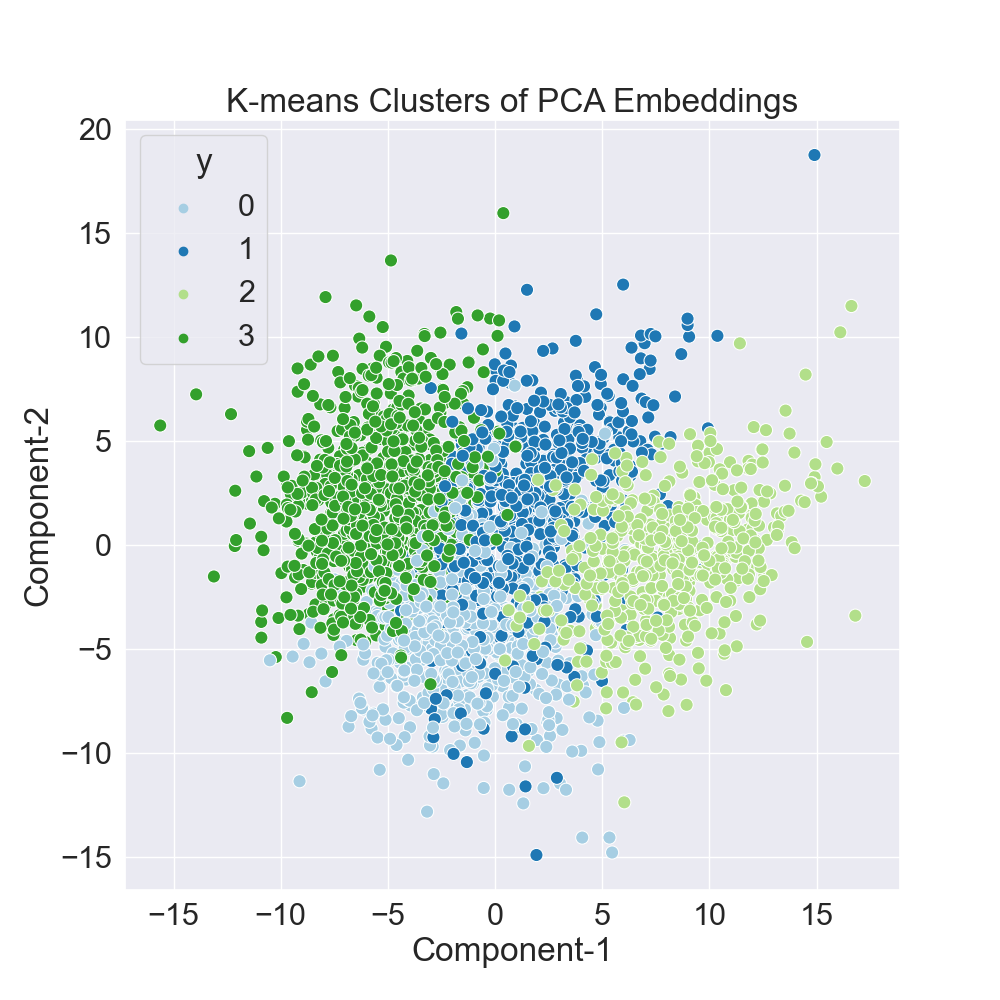}
        \caption{PCA - K-Mean Clusters}
        \label{fig:pca-kmeans}
    \end{subfigure}
    \begin{subfigure}{0.4\textwidth}
    \centering
        \includegraphics[width=5.5cm]{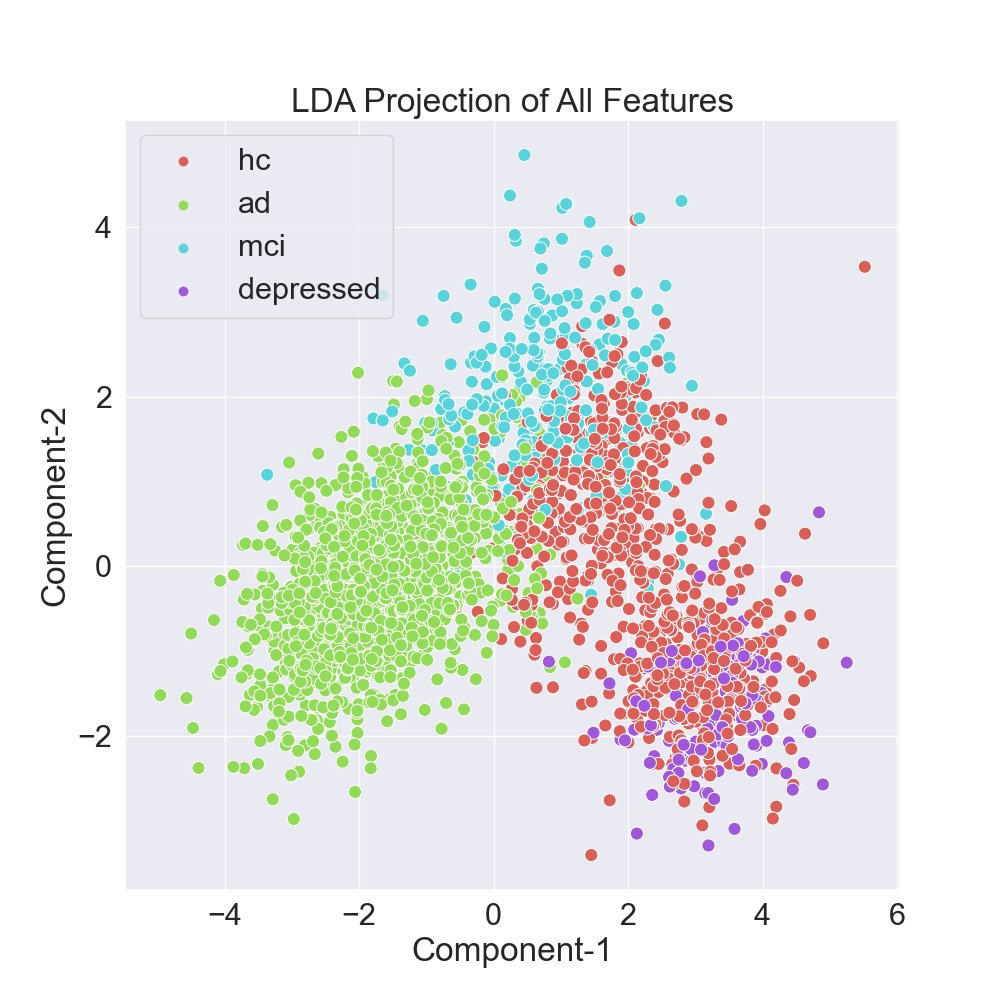}
        \caption{LDA - Disease Clusters}
        \label{fig:lda-disease}
    \end{subfigure}%
    \quad
    \begin{subfigure}{0.4\textwidth}
    \centering
        \includegraphics[width=5.5cm]{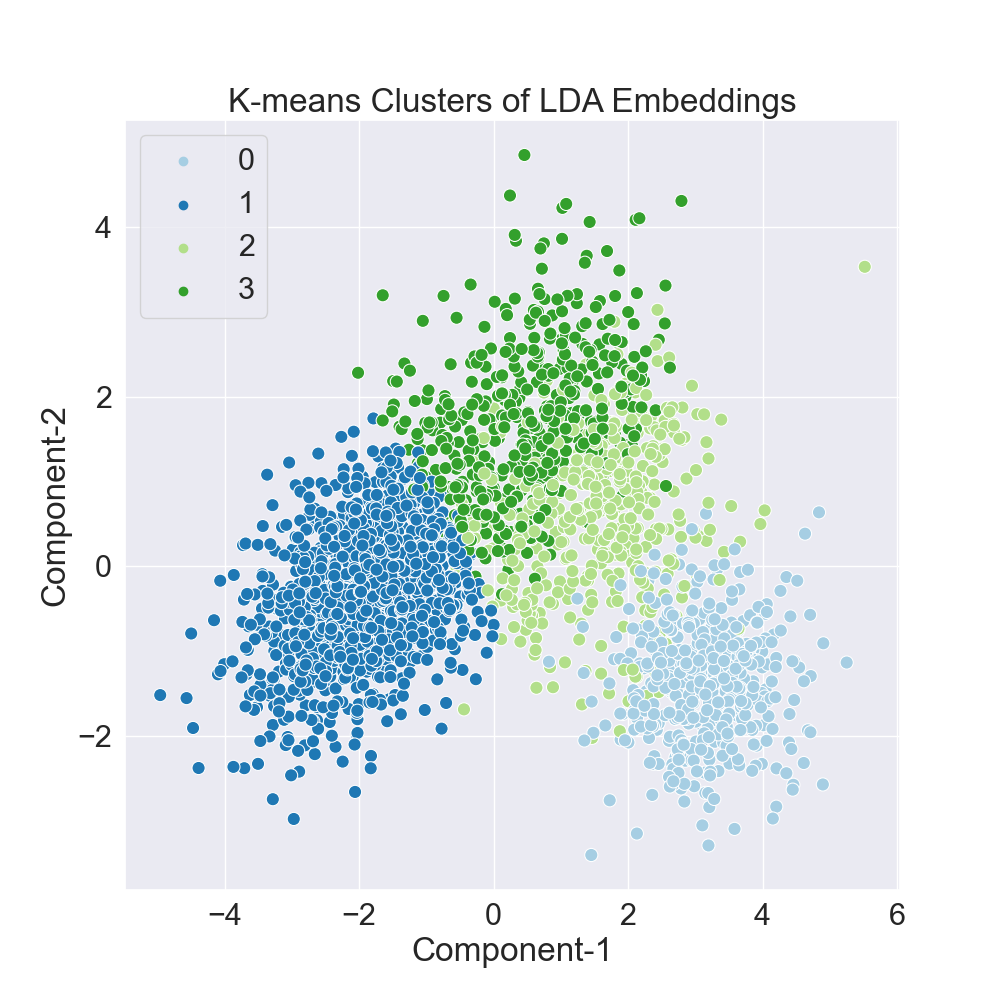}
        \caption{LDA - K-Means Clusters}
        \label{fig:lda-kmeans}
    \end{subfigure}
    \caption{Pairwise scatter plots of the linear dimensionality reduction techniques (Component-1 vs Component-2). Left figures: 2-D representation of the samples colored based on their diagnosis labels. Right figures: 2-D representation of the samples colored based on the data-driven clusters resulted from K-Means clustering for K=4.}
    \label{fig:linear-cluster}
\end{figure*}

\begin{figure*}[t]
    \centering
    \begin{subfigure}{0.4\textwidth}
    \centering
        \includegraphics[width=5.5cm]{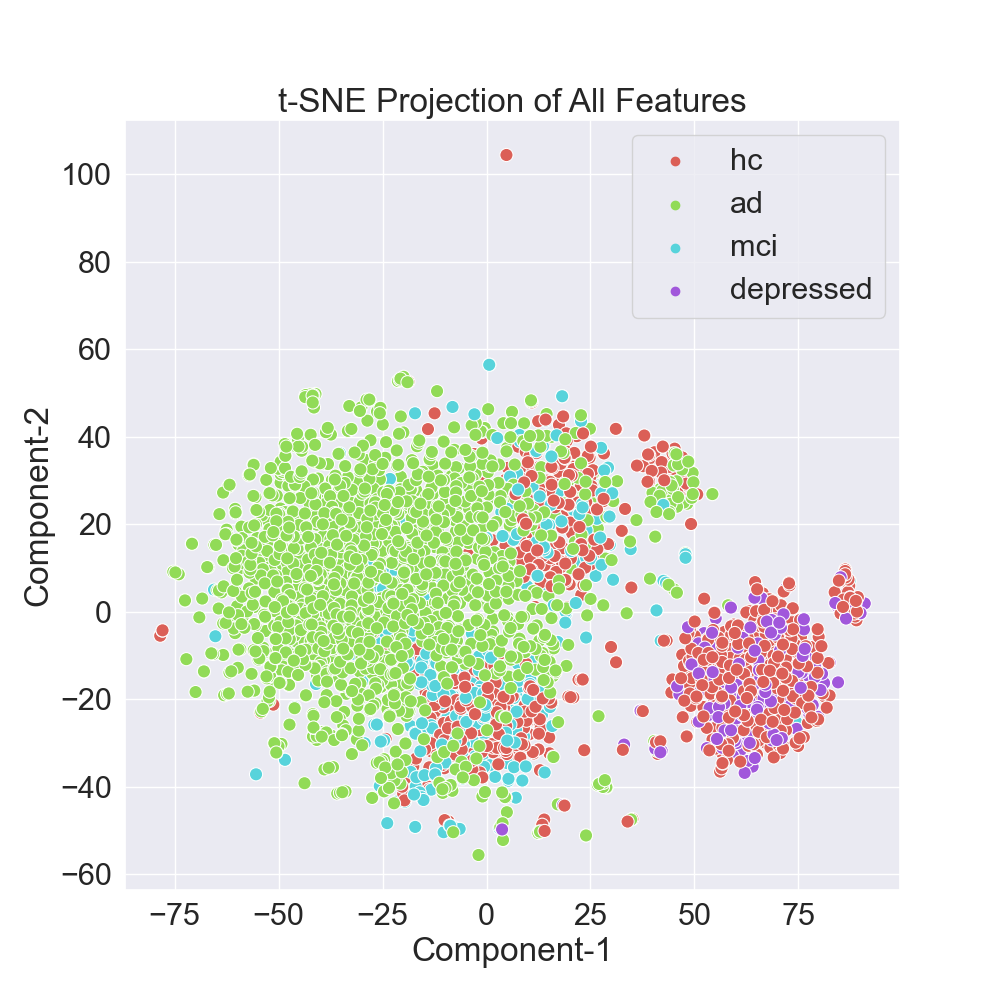}
        \caption{t-SNE - Disease Clusters}
        \label{fig:tsne2-disease}
    \end{subfigure}%
    \quad
    \begin{subfigure}{0.4\textwidth}
    \centering
        \includegraphics[width=5.5cm]{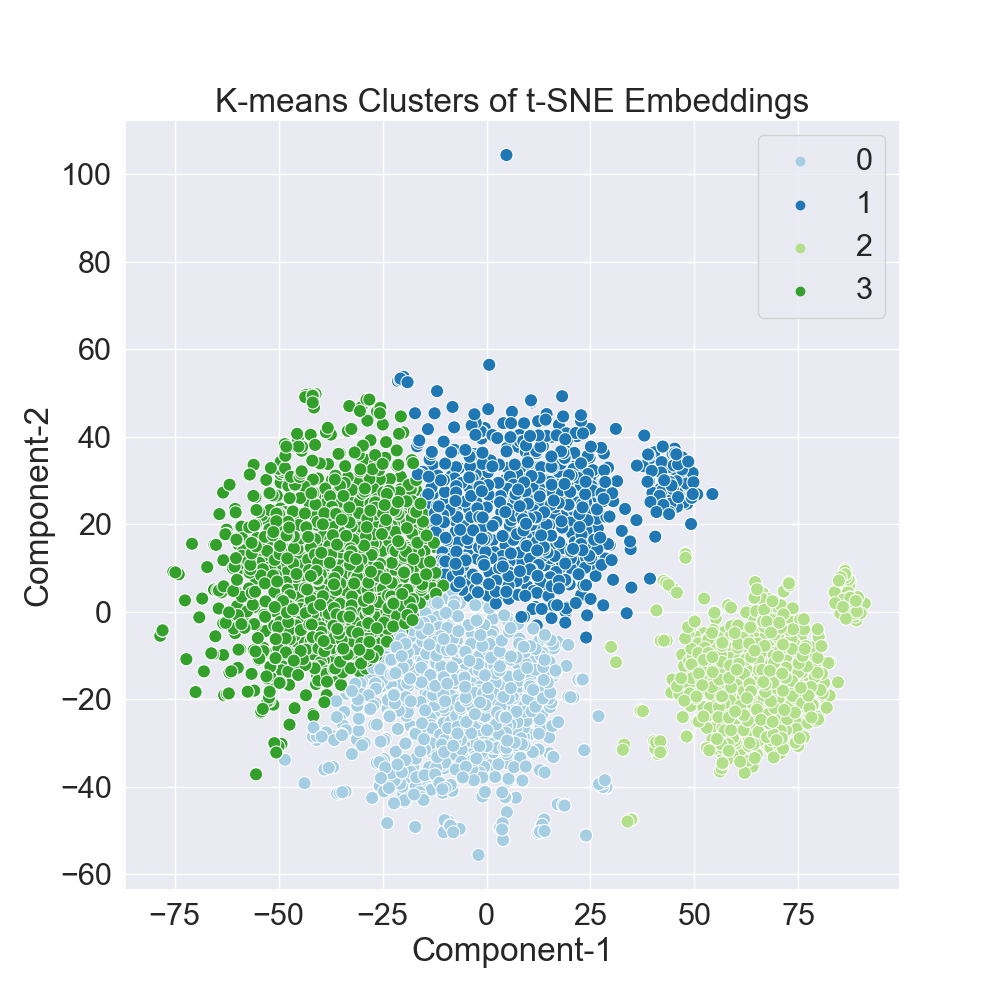}
        \caption{t-SNE - K-Means Clusters}
        \label{fig:tsne2-kmeans}
    \end{subfigure}
    \begin{subfigure}{0.4\textwidth}
    \centering
        \includegraphics[width=5.5cm]{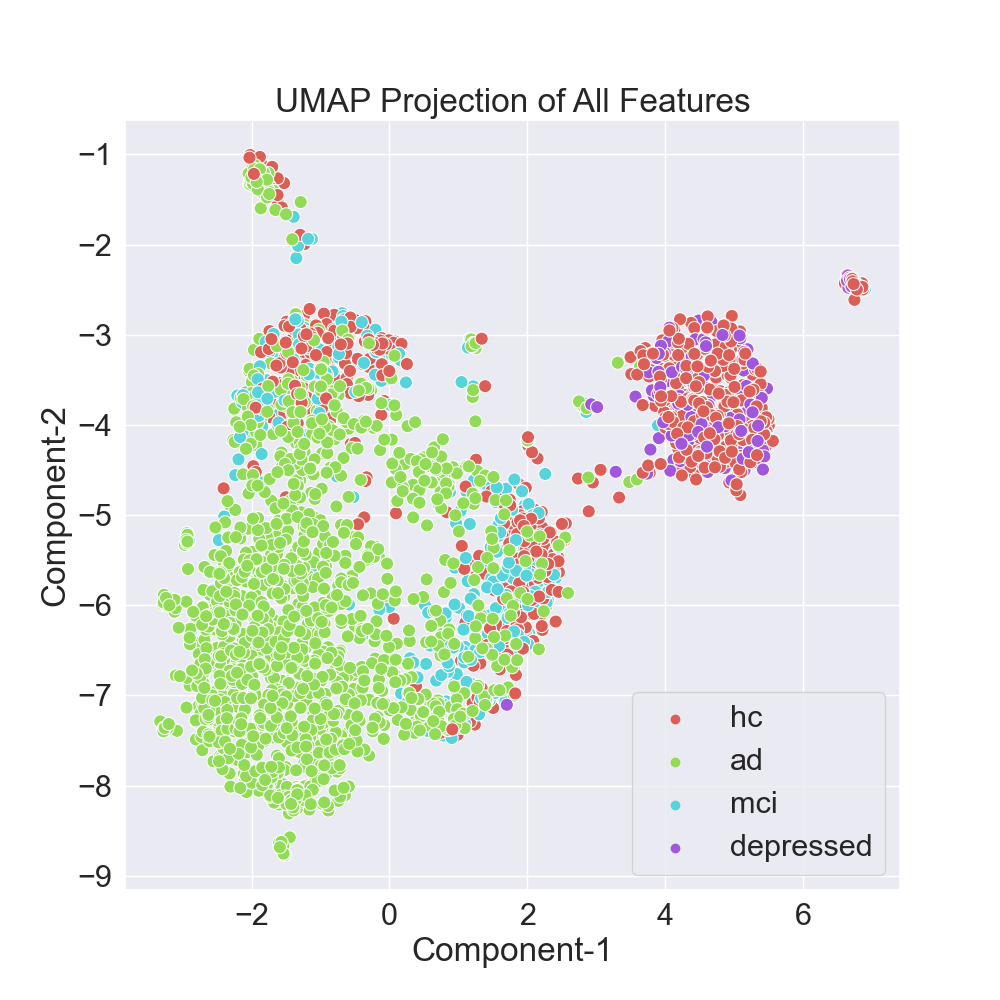}
        \caption{UMAP - Disease Clusters}
        \label{fig:umap-disease}
    \end{subfigure}%
    \quad
    \begin{subfigure}{0.4\textwidth}
    \centering
        \includegraphics[width=5.5cm]{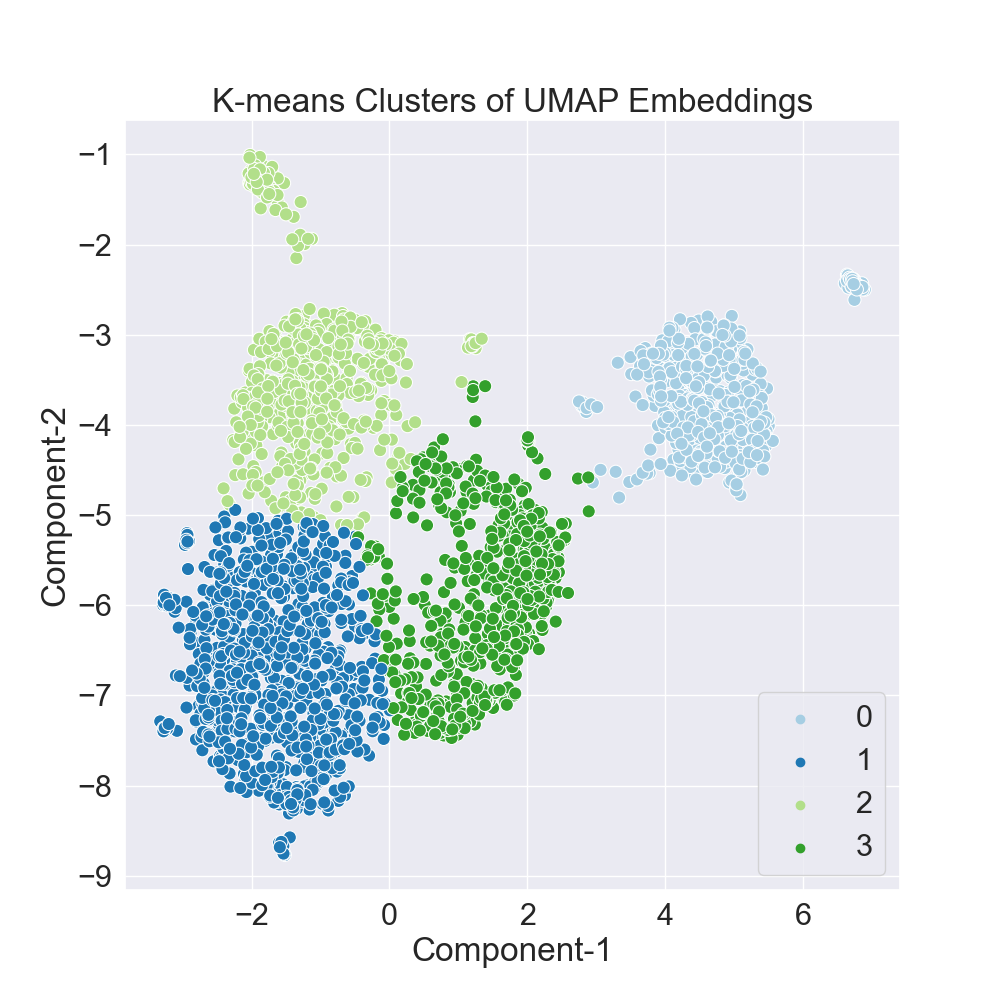}
        \caption{UMAP - K-Means Clusters}
        \label{fig:umap-kmeans}
    \end{subfigure}
    \caption{Pairwise scatter plots of the non-linear dimensionality reduction techniques (Component-1 vs Component-2). Left figures: 2-D representation of the samples colored based on their diagnosis labels. Right figures: 2-D representation of the samples colored based on the data-driven clusters resulted from K-Means clustering for K=4.}
    \label{fig:non-linear-cluster}
\end{figure*}

In this experiment, we investigated what feature categories are the main differentiating factors between the distinguishable disease clusters derived by K-Means clustering. As the first step, we randomly selected 10 HC samples from each cluster and applied Lime-for-t-SNE model to explain the local trends in their neighborhood. We also picked 10 Depr and 10\footnote{We selected 10 samples from each disease cluster, since each group must contain at least 5 samples for both Kruskal-Wallis H-Test \cite{lomuscio2021kruskal} and Mann-Whitney U-Test \cite{bedre2021Mann} explained in \ref{sec:feat-contribution}.} AD data points from the associated disease clusters and followed the same procedure to locally explain the low-dimensional components. Figure \ref{fig:lime-for-tsne} depicts an example of the local explanation of t-SNE embeddings for a selected Depr instance. For each candidate sample, we generated a vector of length 9 indicating the total number of highly-contributed features explaining either quasi-horizontal (e.g., $W1$ in Figure \ref{fig:lime-for-tsne}) or quasi-vertical (e.g., $W2$ in Figure \ref{fig:lime-for-tsne}) trends per each feature category including acoustic, syntactic complexity, discourse mapping, lexical complexity and richness, information content units, sentiment, word finding difficulty, coherence (global and local), and utterance cohesion.

\paragraph{Overall Group Comparison (Kruskal-Wallis H-Test):}
After calculating the feature frequency vectors of the selected samples, we applied overall group comparison per each feature category to test the overall difference between the feature frequencies across the disease groups. For this purpose, we used Kruskal-Wallis H-test \citep{kruskal1952use} using the \verb|scipy.stats.kruskal| library in python. 

\paragraph{Pairwise Group Comparison (Mann-Whitney U-Test):}
As a post-hoc comparison method, we then applied pairwise Mann-Whitney U-test \citep{mann1947test} using the \verb|scipy.stats.mannwhitneyu| python library to determine the distributions of which feature categories are significantly different between each pair of the selected disease groups. 

\subsubsection{Classification of AD vs Depr vs MCI}
After analyzing the feature contributions to the non-linear components, we used the main differentiating feature categories as a feature selection technique to investigate whether they improve the classification performance of AD vs Depr vs MCI. For this purpose, we separately trained Multi-layer Perceptron classifier (MLPClassifier) on the following feature sets:
\begin{enumerate}
  \setlength{\itemsep}{0pt}
  \setlength{\parskip}{0pt}
  \setlength{\parsep}{0pt}
    \item $F$: All the hand-crafted acoustic and linguistic features
    \item $F_d$: Only the feature categories that are shown to be the main differentiators between AD and Depr based on Mann-Whitney test
    \item $F - F_d$: All the hand-crafted features excluding the main distinguishing feature categories
\end{enumerate}

We implemented MLPClassifier by \verb|neural_network.MLPClassifier| package of Scikit-learn \cite{scikit-learn} with all the hyperparameters set to their default parameter settings. We trained the models using grouped 10-fold cross validation to avoid overlapping subjects between the train and test folds and evaluated the performance of the models in terms of macro average accuracy, precision, recall, and F1 scores across the 10 folds.

\section{Results and Discussion}
\subsection{Comparison of Linear and Non-linear Dimensionality Reduction Approaches}
\label{sec:compare-dr}

Table \ref{tab:summary-table} compares how the linear approaches (e.g., PCA, and LDA) perform versus the non-linear techniques (e.g., t-SNE, and UMAP) in distinguishing between distinct diagnosis labels (i.e., AD, MCI, HC, and Depr). Their performance is compared according to their optimal number of K-Means clusters, and Silhouette score. The second column of Table \ref{tab:summary-table} represents whether the optimal number of data-driven disease clusters in K-Means clustering is equal to the total number of diagnosis labels in the aggregated dataset, which is our ideal case.

Between the linear techniques, the Silhouette score obtained by LDA is about twice in value compared to PCA. This can be due to the fact that LDA \citep{izenman2013linear} is a supervised dimensionality reduction technique which focuses on maximizing the class separability by projecting the data points on a new linear axis, while PCA \citep{wold1987principal} tries to find the directions of maximal variance. Based on Figure \ref{fig:lda-disease} and \ref{fig:lda-kmeans}, the clusters of different diseases, as well as the K-Means clusters in LDA, are more visually distinguishable when compared to PCA (See Figure \ref{fig:pca-disease} and \ref{fig:pca-kmeans}). It is also interesting to note how the clusters are placed in LDA plots. MCI comes between AD and HC samples, while depressed data points are positioned on the right end of the figure. This visualization creates a spectrum from AD to MCI, to healthy samples and also, well-separated depressed data points from the rest of the samples.

Interestingly, the optimal number of K-Means clusters in t-SNE is exactly equal with 4 (the total number of disease labels in our data set), which is our ideal case. In addition, its Silhouette score is higher than both PCA and LDA methods. Figure \ref{fig:tsne2-disease} illustrates how well the disease clusters are separated in this model.

In Table \ref{tab:summary-table}, we observe that UMAP demonstrates the best performance among all clustering techniques according to its optimal number of K-Means clusters and Silhouette score. Its optimal number of clusters determined by elbow method is exactly the same as the original number of diagnosis labels. In addition, its Silhoutte score is higher than other approaches meaning that the level of separability of the data-driven disease clusters is higher in UMAP. The associated cluster visualizations for UMAP are also depicted in Figure \ref{fig:umap-disease}. We see depressed samples are well-separated from AD, and MCI, although AD and MCI themselves are not easily distinguishable.

In summary, linear dimensionality reduction techniques like PCA and LDA transform the data to a low-dimensional space as a linear combination of the original variables, while non-linear techniques are applied when the original high-dimensional data contains non-linear relationships \citep{sumithra2015review}. Consequently, our findings suggest that the linearity assumption might be incorrect for our aggregated dataset and hence, this can be another reason why the non-linear dimensionality reduction techniques outperformed the linear ones.
\subsection{Analysis of the Differentiating Feature Categories between the Disease Clusters}
\label{sec:symptom-analysis}
\begin{table*}[t]
    \centering
    \scriptsize
    \begin{tabular}{lccccccccc}
    \hline
    \textbf{\begin{tabular}[c]{@{}l@{}}Compare\end{tabular}} &
      \textbf{Acoustic} &
      \textbf{\begin{tabular}[c]{@{}c@{}}Syntactic\\Complexity \end{tabular}} &
      \textbf{\begin{tabular}[c]{@{}c@{}}Discourse\\Mapping\end{tabular}} &
      \textbf{\begin{tabular}[c]{@{}c@{}}Lexical\\Complexity\\and\\ Richness\end{tabular}} &
      \textbf{\begin{tabular}[c]{@{}c@{}}Information\\Content\\Units\end{tabular}} &
      \textbf{\begin{tabular}[c]{@{}c@{}}Sentiment\end{tabular}}&
      \textbf{\begin{tabular}[c]{@{}c@{}}Word\\Finding\\Difficulty\end{tabular}}&
      \textbf{\begin{tabular}[c]{@{}c@{}}Coherence\\(Global and\\ Local)\end{tabular}}&
      \textbf{\begin{tabular}[c]{@{}c@{}}Utterance\\ Cohesion\end{tabular}} \\ 
    \hline \hline
     AD vs HC& \textbf{x} & - & \textbf{x} & \textbf{x} & \textbf{x} & - & \textbf{x} & \textbf{x} & - \\ 
     Depr vs HC& \textbf{x} & - & - & \textbf{x} & - & \textbf{x} & - & \textbf{x} & - \\
    HC Variations & \textbf{x} & - & - & \textbf{x} & - & \textbf{x} & - & \textbf{x} & - \\
     AD vs Depr& \textbf{x} & - & \textbf{x} & \textbf{x} & - & - & \textbf{x} & \textbf{x} & - \\  
    \end{tabular}
    \caption{Pairwise Mann-Whitney U-Test on frequency vectors of disease groups. For each pair of disease groups, the feature categories with p-value < 0.05 are marked as differentiating symptoms.}
    \label{tab:mann-whitney}
\end{table*}

\begin{table}[htbp]
    \centering
    \scriptsize
      \resizebox{\columnwidth}{!}{
      \begin{tabular}{lccccccccc}
          \hline
          \bfseries Feature Set & \bfseries Precision & \bfseries Recall & \bfseries F1 Score & \bfseries Accuracy\\
          \hline \hline
    $F$ & 0.88 $\pm$ 0.04 & 0.86 $\pm$ 0.04 & 0.87 $\pm$ 0.04 & 0.90 $\pm$ 0.02\\ 
    $F_{d}$  & \textbf{0.90 $\pm$ 0.03} & \textbf{0.88 $\pm$ 0.03} & \textbf{0.89 $\pm$ 0.03} & \textbf{0.92 $\pm$ 0.02} \\ 
    $F-F_{d}$ & 0.74 $\pm$ 0.06 & 0.69 $\pm$ 0.07 & 0.71 $\pm$ 0.06 & 0.82 $\pm$ 0.03\\ \hline
      \end{tabular}
  }
    \caption{Performance of AD vs MCI vs Depr classification using different feature sets. Here, $F$ denotes all hand-crafted acoustic and linguistic features. $F_d$ denotes differentiating feature categories between AD and Depr. $F-F_d$ denotes all features excluding differentiating feature categories.}
    \label{tab:classification-table}
\end{table}

As it is illustrated in Figure \ref{fig:tsne2-kmeans}, K-Means clustering derived four distinct disease clusters in a data-driven way using t-SNE embeddings. Cluster 2 corresponds to the right-most cluster in Figure \ref{fig:tsne2-disease}, which is a mixture of Depr and HC samples. Cluster 3 associates with the AD green clump of points on the left-most side of Figure \ref{fig:tsne2-disease} and clusters 0 and 1 match with the two zones in the middle comprising a combined set of AD, MCI, and HC data points. We randomly selected 10 HC samples from three distinct clusters 0, 1, and 2. We also picked 10 random Depr points from cluster 0 and 10 random AD points from cluster 3. For each instance, we applied LIME-for-t-SNE to explain its local neighbourhood and derive its frequency vector of feature categories (See Section \ref{sec:analyze-feat-cat}). Overall group comparison using Kruskal-Wallis H-Test on the frequency vectors represents that the feature categories including acoustic, lexical complexity and richness, and coherence are significantly different (with p-value < 0.05) across the disease groups including AD, Depr, and different variations of HC. 

As post-hoc group comparison, we applied pairwise Mann-Whitney U-test on each pair of disease groups to assess what feature categories are the main differentiating symptoms across the disease clusters. As it is shown in Table \ref{tab:mann-whitney}, acoustic, lexical complexity and richness, sentiment, and coherence are significantly different across different variations of HC. These differences show variations within the group of healthy samples that can root in the data origin, and data collection procedures. 

Our results indicate that some samples labeled as Depr are similar to HC samples across all the feature categories. This can be due to the distribution of PHQ-9 scores in DEPAC+ dataset with the majority of samples with scores in the range of 5 to 14 from mild to moderate levels of depression severity. Minor levels of depression does not meet the full criteria of major depressive disorder and the symptoms of minor forms of depression are less severe compared to major depressive disorder \cite{shin2021detection}. This increases the risk of confusing modest rates of depression with control samples \cite{cummins2015review}.

Acoustic, discourse mapping (repetitiveness or circularity of speech), lexical complexity and richness, word finding difficulty, and coherence are found to be the main differentiating symptoms between AD and Depr disease clusters. To investigate the effectiveness of our results, we used these feature categories as a feature selection method.

\subsection{Change in Classification Performance}
We reported the performance of classification of AD vs MCI vs Depr in Table \ref{tab:classification-table}. According to paired sample t-test, the expected value of the accuracy, precision, recall, and F1 scores across 10 folds are significantly different between each pair of $F$, $F_d$, and $F-F_d$ feature sets, with p-value < 0.05. Compared to when using all the features, feature selection using only the differentiating feature categories significantly improved the classification performance in terms of all metrics. Also, excluding the differentiating feature categories significantly worsened the performance of the model in classifying the diseases. These observations support that our proposed method shows a promising avenue toward detecting the data-driven symptoms that can successfully differentiate between Depr, AD, and MCI diseases.

\section{Conclusion}
In this work, we generate a novel aggregated dataset composed of a number of speech corpora including a combination of different clinical conditions (e.g., AD, MCI, HC, and Depr). We extract a hand-crafted set of acoustic and linguistic features derived from speech data, which are used as model predictors for discriminating between the diagnosis labels and we categorize these features under data-driven feature categories in line with the clinical symptoms of these diseases. We cluster the samples into distinguishable disease clusters and examine what speech symptoms are the main differentiating factors between the diseases. Based on our findings, non-linear clustering approaches outperform the linear ones in terms of distinguishing between distinct disease clusters. Our results signify that acoustic abnormality, repetitiveness, or circularity of speech, word finding difficulty, coherence, and differences in lexical complexity and richness are the main differentiating symptoms between different types of dementia (e.g., MCI and AD), and depression.

\bibliography{anthology,custom}
\bibliographystyle{acl_natbib}

\clearpage

\appendix
\section{List of the features}
\label{apd:apd_features}
Detailed description of the linguistic and acoustic variables in our conventional feature set is represented respectively in Table \ref{tab:linguistic-features} and Table \ref{tab:acoustic-features}.

\section{Data Preprocessing}
\label{apd:data-preprocessing}
\subsection{Standardization}
In the data preprocessing step, the features with constant values were removed and then, the feature values were standardized by removing the mean and scaling to unit variance. The standard score of a sample $x$ was calculated as:

\begin{equation}
y = \frac{x-\mu}{\sigma}
\end{equation}

here $\mu$ and $\sigma$ are the mean and standard deviation of the sample $x$ in all training samples.

\subsection{Feature selection}
To remove multicollinearity, one of each pair of the features with Pearson correlation higher than 0.9 was removed.

\section{Implementation and Hyperparameter Setting of the Dimensionality Reduction Models}
\label{apd:param-setting}
\subsection{Linear Approaches}
\label{apd:linear-approach}
\paragraph{PCA: }
PCA was implemented by the \verb|sklearn.decomposition.PCA| package in Scikit-learn \citep{scikit-learn} and its number of components was set to the optimal number of Principal Components (PCs) calculated by Horn's parallel analysis \citep{dinno2009implementing}, which was equal to \textbf{46}. After sorting the PCs based on their explained variance ratio, the feature loadings \citep{PCAloadings} were calculated to measure the correlation between the features and the low-dimensional components. According to the distribution of the feature loadings, features with absolute value of loadings \(\geq0.4\) were selected as the highly-correlated features in each PC. The number of components with the largest values of explained variance ratio and at least one highly-correlated feature was chosen as the optimal number of components. As a result, the tuned number of components was equal with \textbf{8}. This approach selects the components which explain the most variance in data and include features which are highly-correlated with PCs on a linear scale.

\paragraph{LDA: } 
LDA was implemented by the \verb|LinearDiscriminimumantAnalysis| package of Scikit-learn \citep{scikit-learn} with its default parameter settings. The number of components was set equal to 3 (the maximum allowed value), which is the number of classes-1, to achieve the highest total explained variance ratio. The classes represent the diagnosis labels in our study including HC, AD, Depr, and MCI.

\subsection{Non-linear Approaches}
\label{apd:non-linear-approach}
\paragraph{t-SNE: } 
t-SNE \citep{van2008visualizing} was implemented by the \verb|sklearn.manifold.TSNE| package of Scikit-learn \citep{scikit-learn}. Perplexity was tuned by grid search to obtain the highest Silhouette score (See Section \ref{sec:performance-metrics}) in K-Means clustering trained on the t-SNE embeddings. The rest of the hyper-parameters were left unchanged with their default values. We used perplexity=30 to preserve both local and global structure of the given data to an adequate level \citep{wattenberg2016how}, in line with the recommended range of perplexity values by \citet{van2008visualizing}. The number of components in t-SNE was manually tuned to 2, which was the best performing one based on the Silhouette score metric (See Section Section \ref{sec:performance-metrics}). 

\paragraph{UMAP: }
This algorithm was implemented using the original \verb|UMAP|\footnote{\href{https://umap-learn.readthedocs.io/en/latest/}{https://umap-learn.readthedocs.io/en/latest/}} library. Among different combinations of parameter settings, grid search indicated that number of components=2, the number of neighbours=50, and minimum distance=0.1 obtained the highest Silhouette score (See Section \ref{sec:performance-metrics}) in K-Means clustering trained on the UMAP embeddings. The remaining parameters were set to their default values.

\onecolumn
\begin{center}
\begin{longtable}{|p{0.25\linewidth}|    p{0.10\linewidth}|p{0.60\linewidth}|}
    \multicolumn{3}{c}{\textbf{Linguistic Features}}\\
    \hline
    \textbf{Feature Category}  &  \textbf{\#Features} &  \textbf{Brief Description}\\
    \hline
    \hline
    \endfirsthead
    Syntactic complexity & 143 & \textbf{Constituency-parsing based features}: Scores based on the parse tree \cite{chae2009predicting} (e.g., the height of the tree, the statistical functions of Yngve depth (a measure of embeddedness) \cite{yngve1960model}, and the frequencies of various production rules\cite{chae2009predicting}).
    
    \textbf{Lu's syntactic complexity features}: Metrics of syntactic complexity suggested by \citet{lu2010automatic} such as the length of sentences, T-units, and clauses, etc.

    \textbf{Utterance length}: Statistical functionals of utterance length.\\
    \hline
    Lexical complexity and richness & 103 &
    \textbf{Grammatical constituents}: The constituents of the parse tree represented in a collection of context-free grammar variables.
    
    \textbf{Vocabulary richness}: Type-token ratios; brunet \cite{brunet1978vocabulaire}; Honore's statistic \cite{honore1979some}.
    
    \textbf{Lexical norm-based}: Average norms across all words, verbs only, and nouns only for imageability, age of acquisition, familiarity \cite{stadthagen2006bristol} and frequency \cite{brysbaert2009moving}.\\
    \hline
    Discourse mapping & 18 & \textbf{Utterance distances} quantifying the utterance similarity via distance metrics and \textbf{speech-graph} \cite{mota2012speech} features based on the graph representation of the transcripts.\\
    \hline
    Global coherence & 15 & Statistical functionals of cosine distance between GloVe \cite{pennington2014glove} word embeddings of each utterance and its nearest content unit centroid utterances.\\
    \hline
    Local coherence & 15 & Statistical functionals of the similarity between Word2Vec \cite{mikolov2013distributed} embeddings of the successive utterances.\\
    \hline
    Word finding difficulty & 11 & \textbf{Pauses and fillers}:
     Variables like hesitation, speech rate, word duration, and number of filled and unfilled pauses as markers of difficulty in finding words resulting in less fluent speech \cite{pope1970anxiety}.
    
    \textbf{Invalid words}: The proportion of words not in the English dictionary (NID).\\
    \hline
    Information units & 10 & The number of information content units including objects, subjects, locations, and actions applied to quantify the number of items correctly named through the picture description task.\\
    \hline
    Sentiment & 9 & Valence, arousal, and dominance scores for all words and word types describing the sentiment of the words used \cite{warriner2013norms}.\\
    \hline
    Utterance cohesion & 1 & Proportion of the number of switches in verb tense across utterances.\\
    \hline
    \caption{List of all hand-curated linguistic features derived from transcripts. The number of features in each feature category is indicated in the second column (titled `\#Features').}
    \label{tab:linguistic-features}
\end{longtable}
\end{center}

\clearpage

\begin{table*}[htbp]
\renewcommand{\arraystretch}{1.3}
    \centering
    \setlength\tabcolsep{2pt}
    \begin{tabular}{|p{0.4\linewidth}|p{0.15\linewidth}|p{0.4\linewidth}|}
    \multicolumn{3}{c}{\textbf{Spectral and Energy Related Features}}\\
    \hline
    \textbf{Feature} & \textbf{\#Features} &  \textbf{Brief Description}\\
    \hline
    \hline
     Mel-Frequency Cepstral Coefficients (MFCC) 0-12 & 168 & Statistical functionals of 42 MFCC coefficients.\\
    \hline
    Intensity & 8 & Statistical functionals of the perceived loudness in $dB$ (auditory model based).\\
    \hline
    Zero-Crossing Rate (ZCR) & 4 & Statistical functionals of zero crossing rate across all the voiced frames.\\
    \hline
    \multicolumn{3}{c}{\textbf{Voicing Related Features}}\\
    \hline
    Harmonic-to-Noise Ratio (HNR) & 12 & Statistical functionals of the degree of acoustic periodicity in dB using both auto-correlation and cross-correlation methods.\\
    \hline
    Jitter and Shimmer & 11 & Jitter indicates the variability or perturbation of fundamental frequency, while shimmer refers to the same perturbation, but it is related to the amplitude of sound wave, or intensity of vocal emission \citep{wertzner2005analysis}.\\
    \hline
    Pauses and Fillers & 8 & Number and duration of short, medium, and long pauses, fillers(um,uh), mean pause duration, and pause-to-speech ratio.\\
    \hline
    Fundamental Frequency $(F_0)$ & 6 & Statistical functionals of the fundamental frequency in Hz.\\
    \hline
    Durational features & 2 & Total sample and speech duration in the audio record.\\
    \hline
    Phonation Rate & 1 & Number of voiced samples over the total number of samples.\\
    \hline
    \end{tabular}
    \caption{List of all hand-curated acoustic features derived from audio records. The number of features in each feature category is indicated in the second column (titled `\#Features').}
    \label{tab:acoustic-features}
\end{table*}

\end{document}